\documentclass[preprint]{jmlr}  

\usepackage{blindtext}
\usepackage{graphicx}
\usepackage{caption}
\usepackage{array}
\usepackage{amsmath}
\usepackage{booktabs}
\usepackage{listings}
\usepackage{amsfonts}
\usepackage{geometry}
\usepackage{float}
\usepackage{color}
\usepackage{xcolor}
\usepackage{multirow}
\usepackage{lipsum} 
\usepackage{enumitem}

\title{A Conceptual Framework For Trie-Augmented Neural Networks (TANNS)}

\author{\Name{Temitayo Adefemi} \\
\Email{T.M.Adefemi@sms.ed.ac.uk} \\
\addr{University of Edinburgh}}

\begin{document}

\maketitle

\begin{abstract}
Trie-Augmented Neural Networks (TANNs) combine trie structures with neural networks, forming a hierarchical design that enhances decision-making transparency and efficiency in machine learning. This paper investigates the use of TANNs for text and document classification, applying Recurrent Neural Networks (RNNs) and Feed forward Neural Networks (FNNs). We evaluated TANNs on the 20 NewsGroup and SMS Spam Collection datasets, comparing their performance with traditional RNN and FFN Networks with and without dropout regularization. The results show that TANNs achieve similar or slightly better performance in text classification. The primary advantage of TANNs is their structured decision-making process, which improves interpretability. We discuss implementation challenges and practical limitations. Future work will aim to refine the TANNs architecture for more complex classification tasks.
\end{abstract}

\section{Introduction}

Machine learning has significantly evolved, transitioning from foundational architectures such as recurrent and convolutional neural networks to advanced models like the Transformer, which underpins language models like the GPT series. These developments have bolstered capabilities in natural language processing and image recognition. Despite these advancements, challenges in interpretability and scalability remain, which limit broader applications.

This paper introduces Trie-Augmented Neural Networks (TANNs), a novel framework that integrates trie data structures with neural networks to improve transparency and learning efficiency. TANNs incorporate embedded neural networks within each trie node, facilitating incremental and interpretable decision-making. This design allows for adaptive learning tailored to specific characteristics of the input, addressing critical limitations of contemporary neural architectures.

We provide a conceptual foundation for TANNs and demonstrate their application in addressing machine learning challenges, particularly in text classification tasks. This framework sets a conceptual groundwork for future research, promoting further exploration into optimizing training and configurations of the TANN architecture, without the need for extensive experimental validation at each step.

\section{Related Work}
Trie-augmented Neural Networks (TANNs) draw
from the established successes of traditional neural
network models like CNNs and RNNs, which excel
in image recognition and natural language processing,
respectively (LeCun et al., 1998; Mikolov et al., 2010).
Despite their effectiveness, these models often need
help with interpretability and scalability.
TANNs build on using hierarchical structures to
improve machine learning models. This principle is
seen in decision trees and random forests, which use
hierarchical data partitioning to enhance decisionmaking and transparency (Quinlan, 1986; Breiman,
2001). This approach is also evident in Hierarchical
Recurrent Neural Networks, which process sequential
data to capture long-term dependencies (El et al.,
1996).
Further, modular neural network advancements, such
as the Mixture of Experts, demonstrate the benefits
of specialized subnetworks for different segments of
the input space, boosting efficiency and performance
(Shazeer et al., 2017). The TANNs architecture
extends this by integrating neural networks within a
trie structure, combining hierarchical organization
with neural adaptability.
Moreover, hybrid models like Neural Turing
Machines, which combine neural networks with
external memory, showcase the potential of neural
networks to manage complex data structures (Graves
et al., 2014). TANNs align with this innovation,
aiming to enhance interpretability and efficiency by
merging neural networks with trie structures.
Thus, TANNs contribute to the evolution of neural
network architectures by incorporating hierarchical
and modular strategies, providing a novel approach to
tackling scalability and interpretability challenges in
machine learning.

\section{Model Architecture}

\subsection{Formal Definition of Trie-Augmented Neural Networks (TANNs)}
\begin{figure}[htbp]
    \centering
    \includegraphics[width=0.4\textwidth]{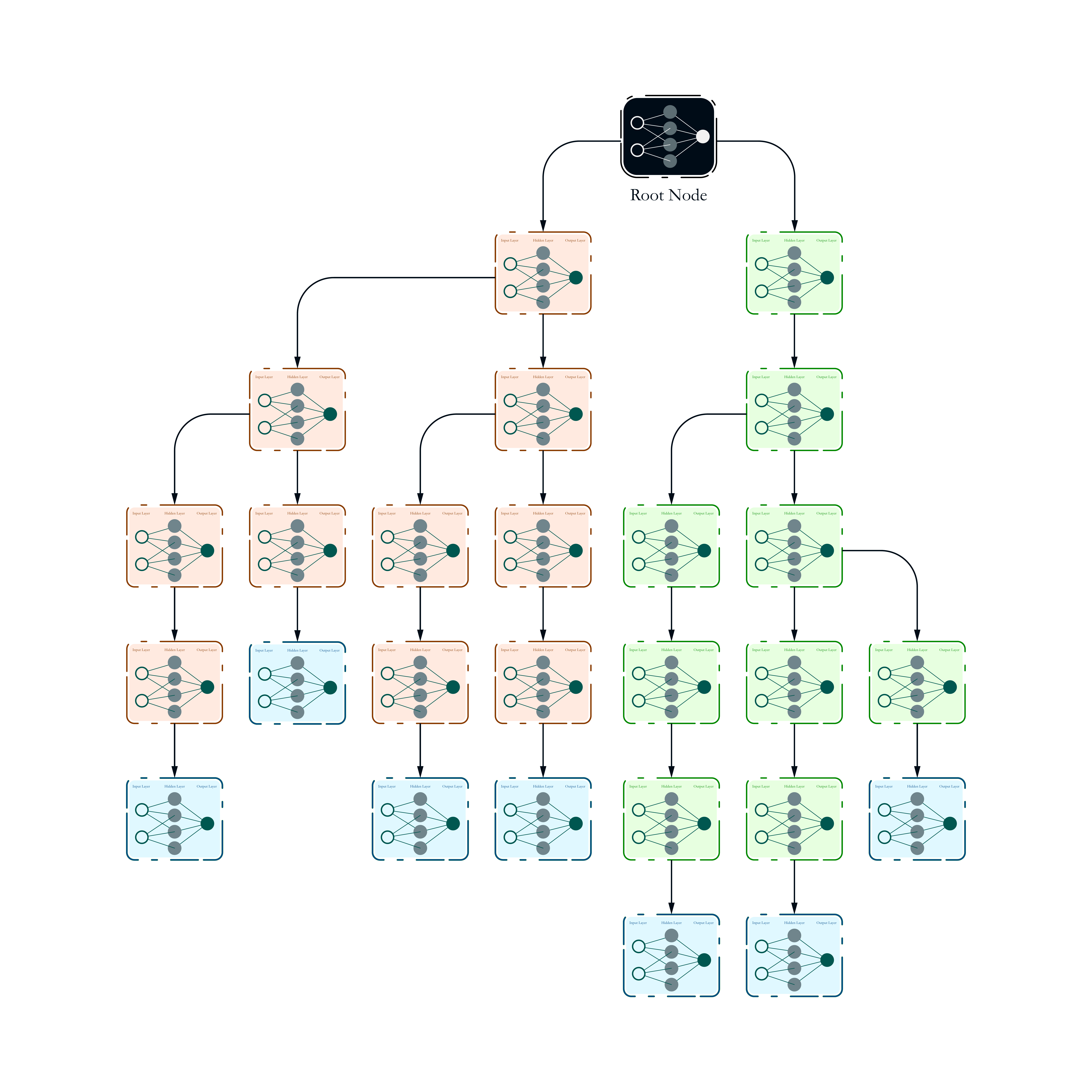}
    \caption{Illustration of the TANN architecture}
    \label{fig:TANN_architecture}
\end{figure}

This section outlines the mathematical structure of Trie-Augmented Neural Networks (TANNs), an architecture that melds the hierarchical layout of tries with neural networks' computational capabilities. The primary goal of TANNs is to efficiently segment the input space and enhance the interpretability and scalability of neural network decisions.

\vspace{0.5cm} 

\textbf{Definition:} A Trie-Augmented Neural Network (TANN) can be defined as a directed acyclic graph \( T = (N, E) \), where:
\begin{itemize}
    \item \( N \) represents the set of nodes within the trie. Each node in this set is not merely a passive point of data routing but actively incorporates a neural network that processes and transforms the data passing through it. Each neural network can be tailored to the specific data characteristics expected at that point in the trie, allowing for specialized processing and learning capabilities that are contextually relevant.
    \item \( E \) represents the set of edges that connect the nodes in \( N \). Each edge in \( E \) defines a hierarchical relationship between nodes, directing data flow from one node to another. This structure ensures that data is processed stepwise, hierarchically, with each node's output as input for one or more subsequent nodes in the trie.
\end{itemize}

\begin{figure}[h!]
    \centering
    \includegraphics[width=0.6\textwidth]{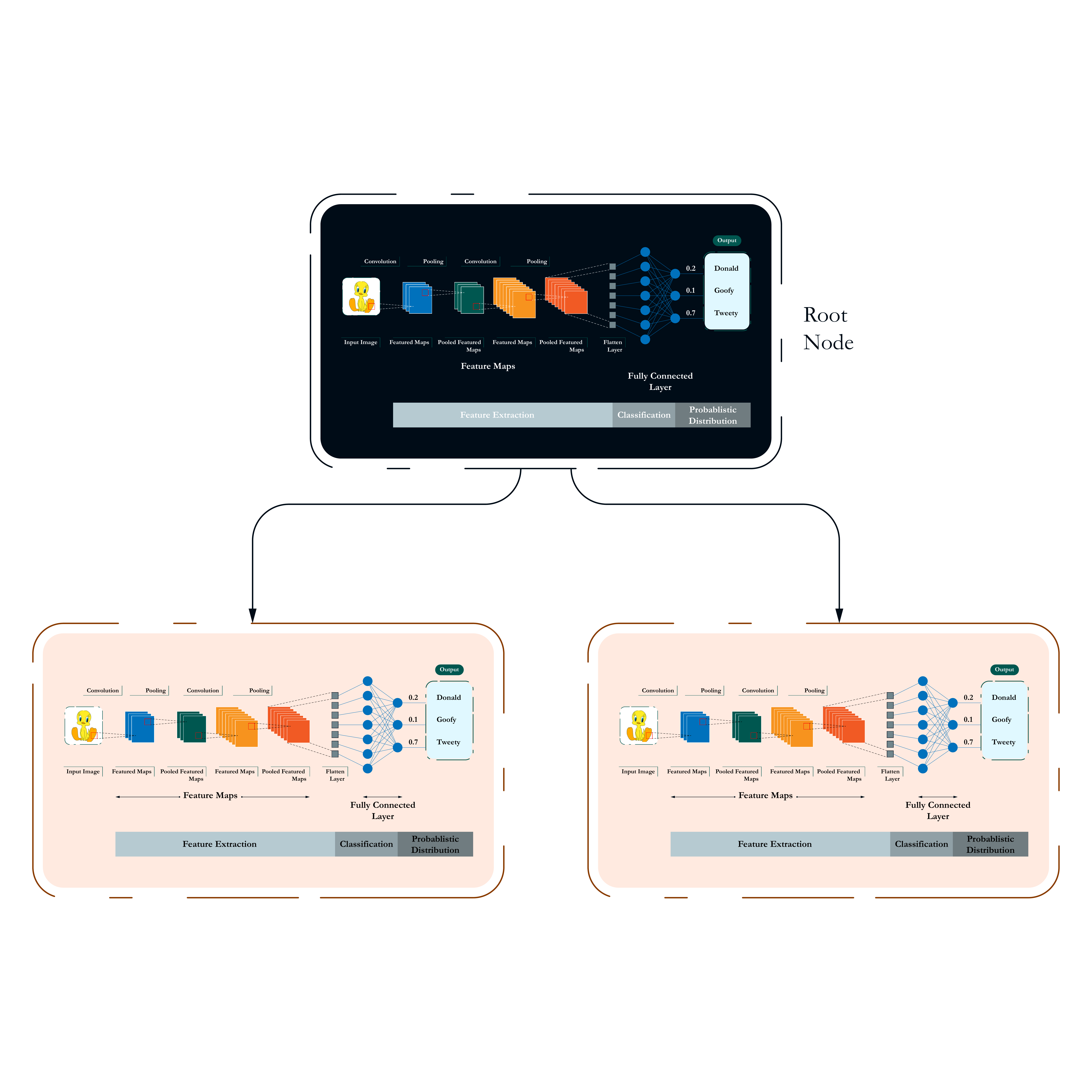}
    \caption{Convolutional Neural Network in Trie-Augmented Neural Network}
    \label{fig:CNN_in_TANN}
\end{figure}

TANNs are network-agnostic, supporting various neural networks like Convolutional Neural Networks (CNNs), Feedforward Neural Networks (FNNs), and Recurrent Neural Networks (RNNs), depending on the task requirements at each node. This flexibility allows each node to address specific aspects of the problem, enhancing the system's overall robustness by isolating input effects and facilitating independent optimization of each module. For instance, CNNs might be employed at nodes dealing with spatial data, while RNNs manage sequential data, optimizing processing across diverse data types.

The hierarchical, modular design of TANNs naturally supports complex decision-making tasks that benefit from breaking down the problem into smaller, manageable segments, enabling more precise and scalable solutions.

\begin{figure}[h!]
    \centering
    \includegraphics[width=0.5\textwidth]{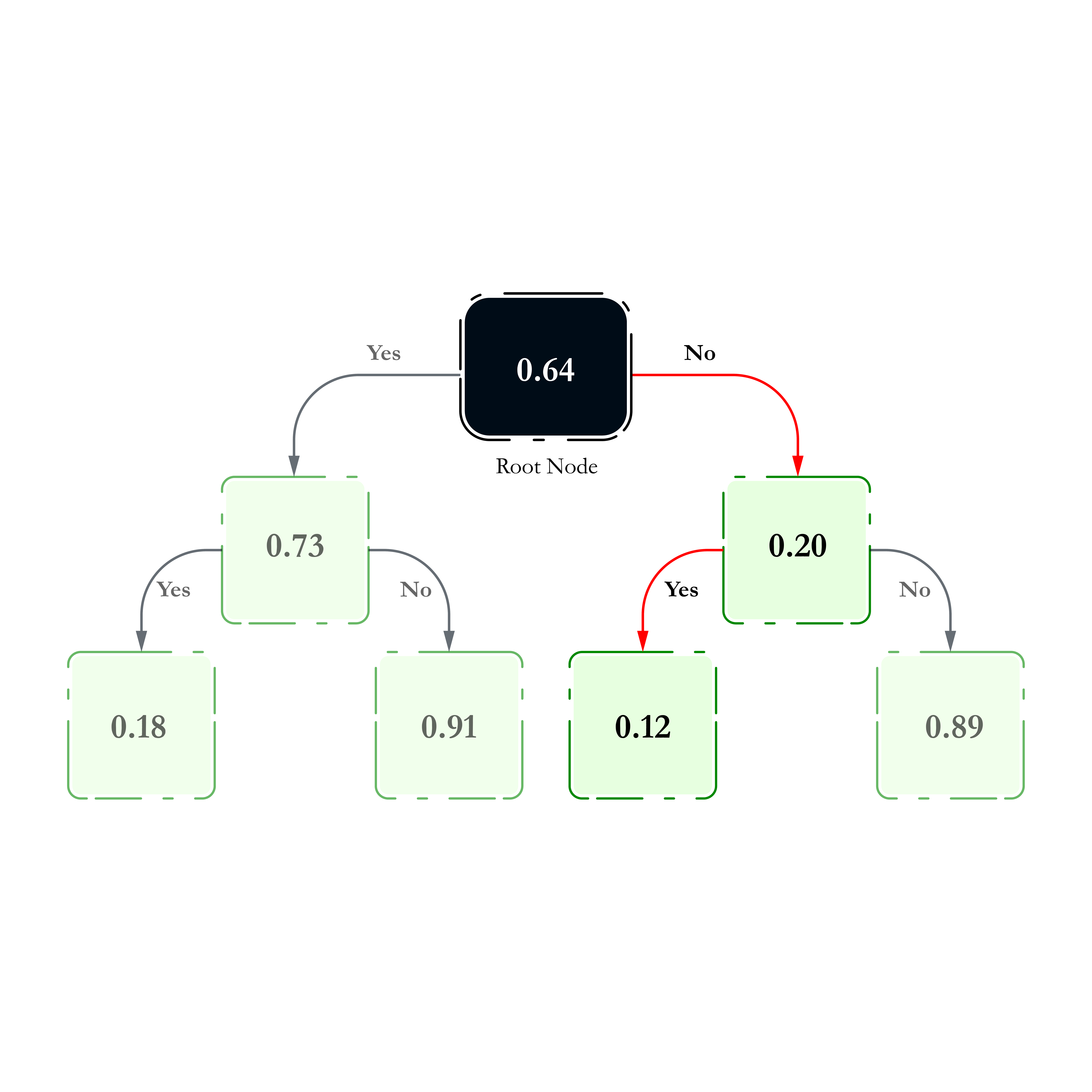}
    \caption{TANN Decision-Making}
    \label{fig:TANN_decision_making}
\end{figure}

\section{Validity}

\subsection{XOR Problem}
We tested the Trie Augmented Neural Network (TANN) on the classic XOR benchmark problem over ten epochs using a balanced Trie depth of 3 with fixed hyperparameters to validate the architecture. The XOR (exclusive OR) problem is a well-known challenge in neural networks, requiring the network to learn a non-linearly separable function.

\vspace{0.5cm} 
The XOR dataset consists of four input patterns:
\begin{itemize}
    \item [0, 0] $\Rightarrow$ 0
    \item [0, 1] $\Rightarrow$ 1
    \item [1, 0] $\Rightarrow$ 1
    \item [1, 1] $\Rightarrow$ 0
\end{itemize}

\vspace{0.2cm} 

\subsection{AND/OR Logic gates}
To further validate the Trie Augmented Neural Network (TANN) and to underscore the learning capabilities before refining and iterating the program to solve more significant problems, we tested the TANN on the classic AND/OR Logic gates benchmark problem over the same ten epochs using a balanced Trie depth of 3 with fixed hyperparameters for both programs to further validate the architecture. The dataset consists of 2 sets with four input patterns each:

\vspace{0.2cm} 
The first set is the AND dataset, which consists of:
\begin{itemize}
    \item [0, 0] $\Rightarrow$ 0
    \item [0, 1] $\Rightarrow$ 0
    \item [1, 0] $\Rightarrow$ 0
    \item [1, 1] $\Rightarrow$ 1
\end{itemize}

\vspace{0.2cm} 
The second set is the OR dataset, which consists of:
\begin{itemize}
    \item [0, 0] $\Rightarrow$ 0
    \item [0, 1] $\Rightarrow$ 1
    \item [1, 0] $\Rightarrow$ 1
    \item [1, 1] $\Rightarrow$ 1
\end{itemize}

\vspace{0.5cm} 

\begin{table}[h!]
    \centering
     \begin{tabular}{lccccc}
        \hline
        \textbf{Parameter} & \textbf{Value} \\
        \hline
        input\_size & 2 \\
        \hline
        hidden\_size & 20 \\
        \hline
        epochs & 10 \\
        \hline
    \end{tabular}
    \caption{Hyperparameters Table}
    \label{fig:hyperparameters}
\end{table}

\section{Validity Results}
\begin{figure}[H]
    \centering
    \includegraphics[width=0.8\textwidth]{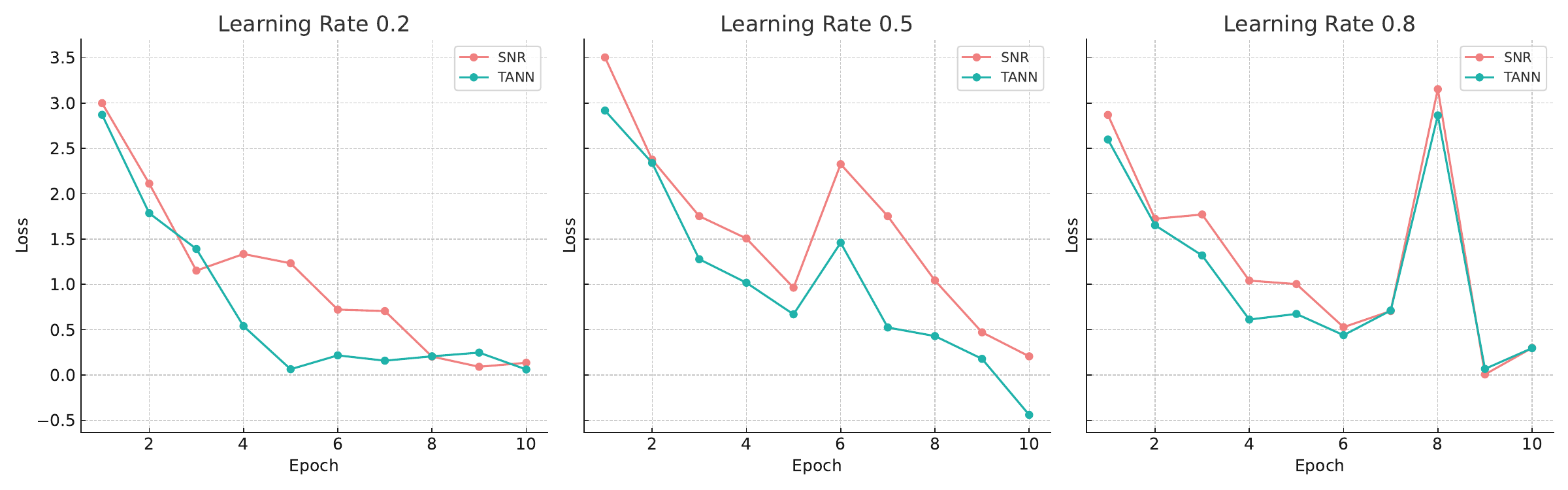}
    \caption{Training Loss for the XOR Benchmark}
    \label{fig:TANN_decision_making}
\end{figure}

\begin{figure}[H]
    \centering
    \includegraphics[width=0.8\textwidth]{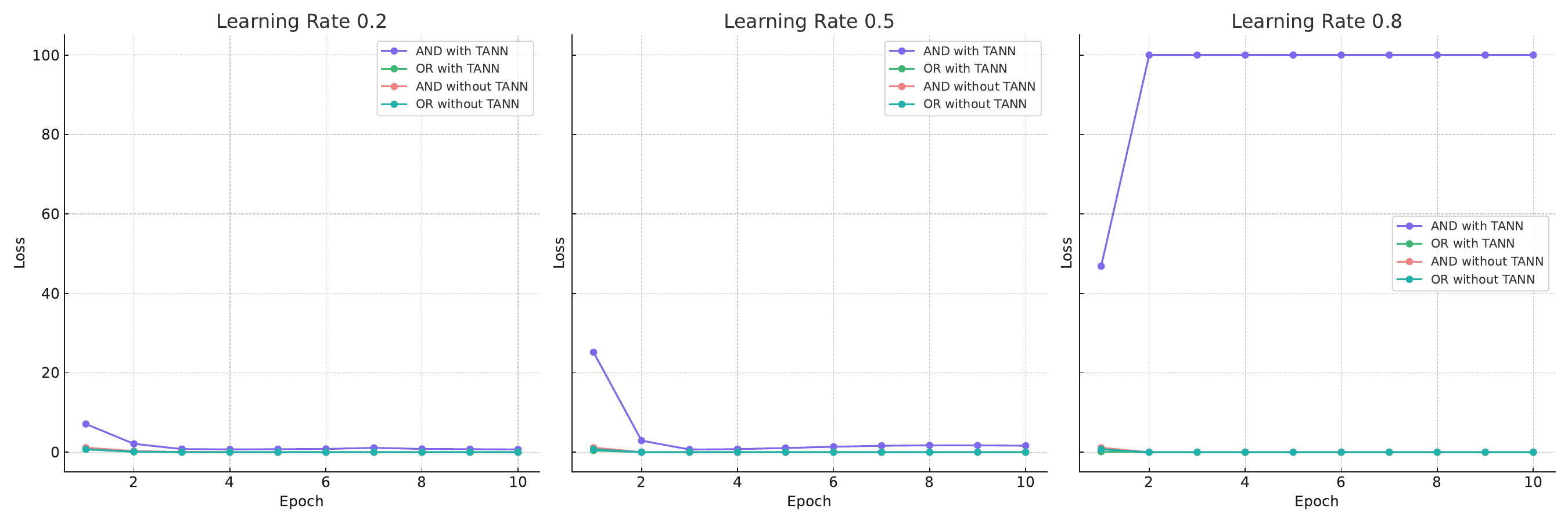}
    \caption{Training Loss for the AND/OR Benchmark}
    \label{fig:TANN_decision_making}
\end{figure}

\vspace{0.5cm} 

In experiments involving the XOR problem—a benchmark for assessing a model's ability to learn nonlinear patterns—the TANN architecture demonstrated superior learning speeds and convergence compared to traditional neural networks. It managed to learn the underlying patterns more effectively, as evidenced by lower final loss values. This indicates a robust capability of the TANN to capture complex data relationships efficiently.

Further testing using AND/OR logic gates emphasized the effectiveness of the TANN. Notably, while changes in the learning rate significantly affected the performance of traditional neural networks, the TANN exhibited stable performance, with minimal variation in loss rates. This consistency and reliability highlight the TANN’s potential for applications in diverse settings.

These positive outcomes strongly support the design of the TANN and its suitability for complex, large-scale applications. The architecture's demonstrated ability to efficiently learn and generalize from intricate patterns suggests it could be extremely useful in advanced tasks involving pattern recognition, classification, and decision-making.

Our ongoing efforts to refine this architecture are aimed at maximizing the TANN’s potential to address large-scale challenges across various domains, potentially setting a new standard in neural network capabilities.

Further investigations included conducting experiments at various trie depths to better understand how the depth of the trie affects the model's loss rate.
\vspace{0.2cm}

The results of these experiments are illustrated in the graph below:

\begin{figure}[H]
    \centering
    \includegraphics[width=0.7\textwidth]{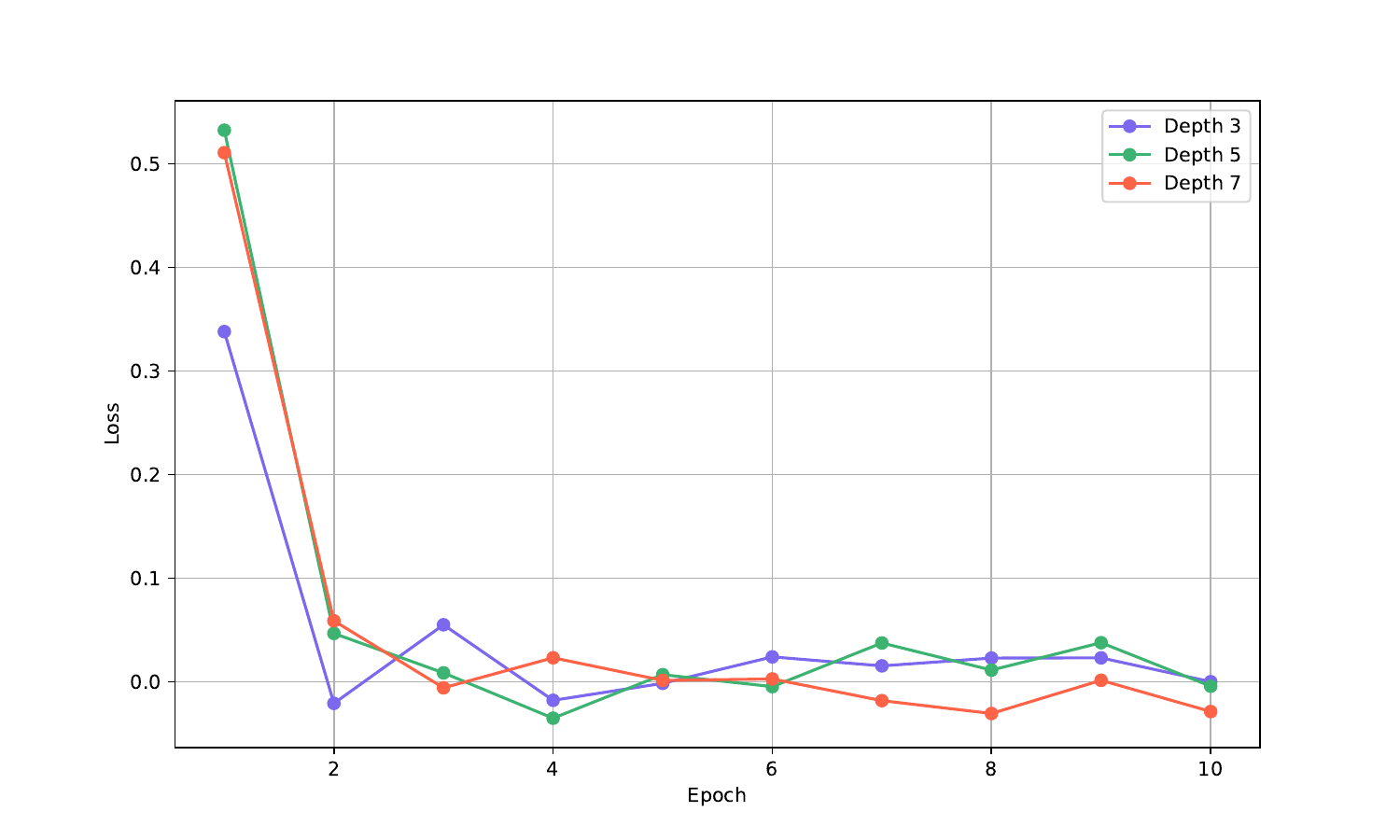}
    \caption{XOR Results across various depths}
    \label{fig:TANN_decision_making}
\end{figure}

\vspace{0.5cm} 

Based on the experimental results, it is evident that the depth of the trie does not significantly impact the training performance of a Trie Augmented Neural Network (TANN) for the given problem. However, it is essential to note that the dataset used in this study is relatively simple and plain. To fully understand the impact of trie depth on performance, further experimentation with more complex and larger datasets is necessary.

\vspace{0.5cm} 
The following section will explore how TANNs can be integrated into standard neural network architectures. We will compare the performance of systems with and without TANN implementations to demonstrate the impact of this architecture.

\subsection{Comparison With Neural Network Architectures}

When addressing the XOR and AND/OR problems, it is critical to assess neural networks alongside other methodologies. Unlike binary classification approaches that employ automata theory, such as decision trees and Tsetlin machines, neural networks are adept at learning intricate patterns within the data. 

\vspace{0.5cm} 

In our studies, we explored several neural network architectures, including complex-valued neurons, convolutional neural networks, recurrent neural networks, and simple neural networks enhanced with dropout regularization. We developed two variants of each architecture: one standard model and one augmented with the Trie-Augmented Neural Network (TANN) architecture, where neural networks are embedded in each node of a balanced trie with a depth of three.

\vspace{0.5cm} 

This research aims to ascertain the efficacy of the Trie-Augmented Neural Network (TANN) in enhancing neural network learning. Experimental outcomes reveal that TANN significantly mitigates training loss compared to conventional neural networks under the same hyperparameters. This enhancement is attributed to the TANN's proficiency in efficiently partitioning the input space through a trie data structure, which facilitates more effective data representation and processing. While initial experiments on foundational benchmarks like the XOR and AND/OR logic gates yield promising results, there is a need for further studies to evaluate the TANN’s performance on more complex and extensive datasets.

\vspace{1cm} 

\begin{table}[H]
    \centering
    \begin{tabular}{lccccc}
        \hline
        \textbf{Model} & \textbf{Learning Rate} & \textbf{Optimizer} & \textbf{Loss Function} & \textbf{Epochs} & \textbf{Dropout Rate} \\
        \hline
        SNN w Dropout & 0.2 & SGD & BCELoss & 10 & 50\% \\
        \hline
        CNN & 0.2 & SGD & BCELoss & 10 & None \\
        \hline
        RNN & 0.2 & SGD & BCELoss & 10 & None \\
        \hline
        CNN & 0.2 & Adam & MSELoss & 10 & None \\
        \hline
    \end{tabular}
    \caption{Basic Model Configurations}
    \label{tab:basic_model_configurations}
\end{table}

\begin{table}[H]
    \centering
    \begin{tabular}{lccccc}
        \textbf{Model} & \textbf{Hidden Layers} & \textbf{Neurons Per Layer} & \textbf{Additional Features} \\
        \hline
        SNN w Dropout & 1 & 4 (first layer) & Sigmoid activation \\
        \hline
        CNN & Conv1d + FC & 1 (each layer) & 1D convolution, Sigmoid \\
        \hline
        RNN & RNN layer & 2 (hidden size) & Batch-first RNN, Sigmoid \\
        \hline
        CNN & 2 & 2 (first layer) & Complex values, ReLU, Magnitude output \\
        \hline
    \end{tabular}
    \caption{Model Structure \& Features}
    \label{tab:model_structure_features}
\end{table}
\vspace{0.5cm} 

The graph below illustrates the results of our experiments:

\begin{figure}[h!]
    \centering
    \includegraphics[width=0.7\textwidth]{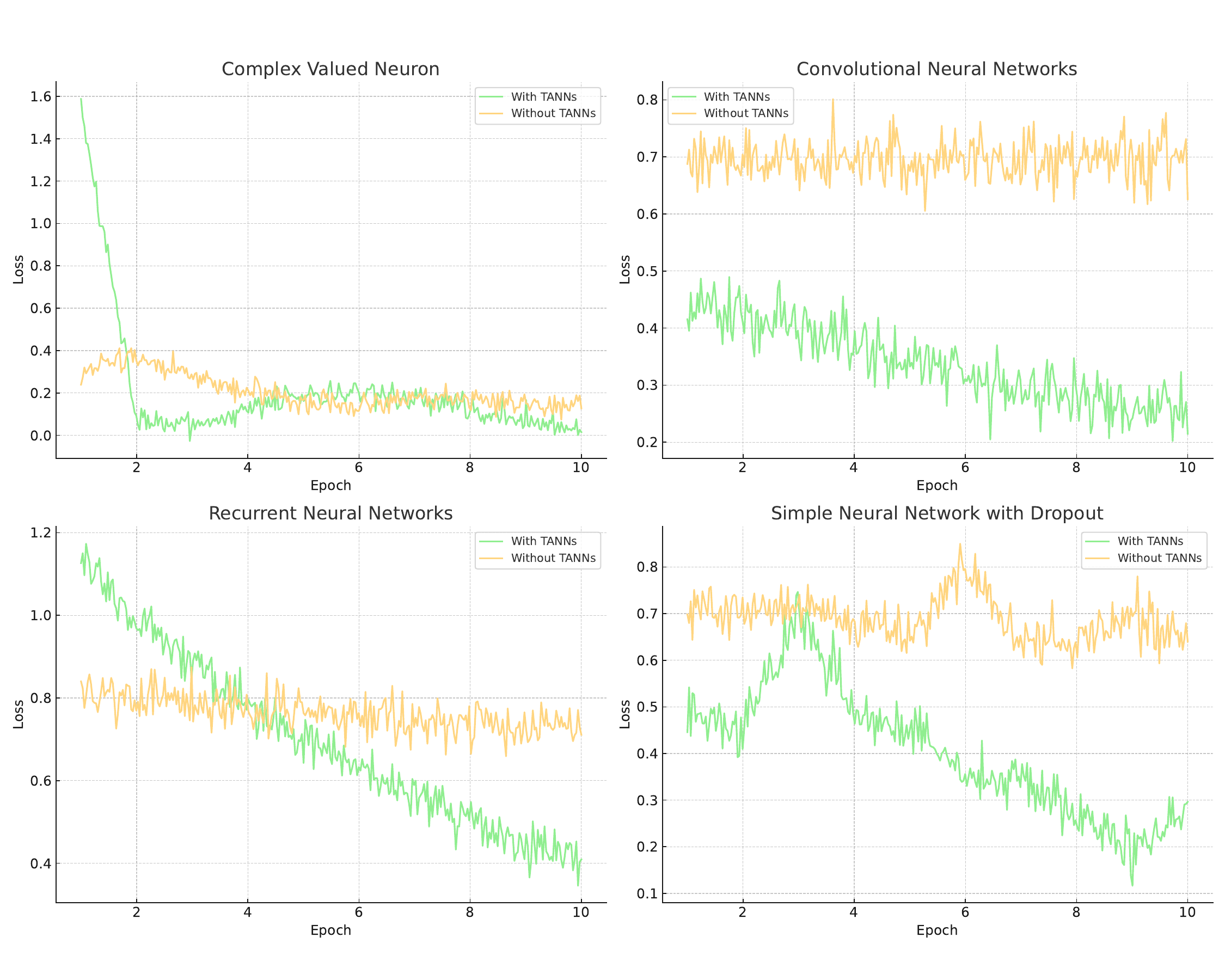}
    \caption{Training loss progression over 10 epochs for four neural network models, with and without Trie-Augmented Neural Networks (TANNs).}
    \label{fig:TANN_decision_making}
\end{figure}

\section{Case Study}

This research paper evaluates the predictive capabilities of the Trie Augmented Neural Network (TANN) architecture specifically within the realm of text-based applications. We focus on document classification using the 20 Newsgroup dataset and spam identification using the SMS Spam Collection Dataset. By concentrating on these textual data domains—structured text and unstructured text respectively—we aim to assess the adaptability and performance of TANNs in handling diverse text classification challenges. This targeted approach allows us to thoroughly investigate the potential of TANNs to enhance interpretability and efficiency in text-based machine learning tasks, providing an in-depth evaluation of their effectiveness in specific scenarios rather than a broad generalizability across unrelated domains.

\vspace{0.1cm} 

Our analysis will include comparisons with two neural network architectures. Document classification and text identification tasks with the SMS Spam Collection Dataset will be analyzed using Recurrent Neural Networks (RNNs) and Feed Forward Neural Networks (FNNs).

\subsection{Training Procedure for Trie-Augmented Neural Networks (TANNs) in this Study}

The training methodology for Trie-Augmented Neural Networks (TANNs) implemented in this study encompasses a meticulously structured sequence of steps aimed at optimizing the learning outcomes. The training initiates with the definition of a loss function using the cross-entropy criterion, represented as \(\mathcal{L} = \text{CrossEntropyLoss}()\). This function quantifies the discrepancy between predicted classifications and actual labels, serving as the pivotal metric for training evaluation.

\vspace{0.1cm} 
Simultaneously, we employ Adam optimizers for each mini neural network within the trie structure, formulated as \(\text{optim.Adam}(\theta, \eta = 0.001)\), where \(\theta\) represents the adjustable parameters of each mini neural network and \(\eta\) signifies the learning rate. The training process progresses over a predefined number of epochs, denoted by \(E\). At the onset of each epoch, the cumulative loss \(\mathcal{L}_{\text{running}}\) is reset to zero, and a batch counter \(C\) is similarly initialized.

\vspace{0.1cm} 
During each training epoch, batches of input data and target labels, represented as \(X\) and \(Y\) respectively, are retrieved from the data loader. For each batch, the batch size \(B\) is ascertained, and an output tensor \(O\), dimensioned as \(B \times \text{num\_classes}\), is initialized to zero. 

\vspace{0.1cm} 
The node traversal within the trie is determined by specific feature values from each input tensor \(x_i\). Specifically, the feature value at the current node's designated feature index, denoted as \(f = x_i[\text{node.feature\_index}]\), guides the traversal direction: if \(f < 0.5\), the traversal proceeds to the left child; otherwise, it advances to the right child. This decision-making process is iteratively conducted until a leaf node is reached. The mini neural network housed at the leaf node then processes \(x_i\), producing the corresponding output. This systematic approach ensures that each input is optimally classified based on the hierarchical structure of the trie, leading to an effective training session.

\subsection{The Datasets}

\subsubsection{20 Newsgroups}

The 20 Newsgroups dataset is a collection of approximately 20,000 newsgroup documents, partitioned nearly evenly across 20 different newsgroups. Each newsgroup corresponds to a distinct topic. Some newsgroups are closely related (e.g., comp.sys.ibm.pc.hardware and comp.sys.mac.hardware), while others are highly unrelated (e.g., misc.forsale and soc.religion.christian). Table 4 specifies the 20 Newsgroups categories and their sizes.

\subsubsection{SMS Spam Collection}

The SMS Spam Collection Dataset is a widely-used dataset for research in the fields of text mining and natural language processing, specifically targeting spam detection. It comprises 5,574 SMS messages, each labeled as either "spam" or "ham" (non-spam). Created by Almeida, Hidalgo, and Yamakami in 2011, this dataset is known for its balance and diversity in message content, making it an ideal resource for training and evaluating machine learning models. The dataset includes various types of spam messages, such as advertisements and phishing attempts, as well as regular, non-spam messages, providing a comprehensive ground for developing and testing spam detection algorithms. Researchers leverage this dataset to explore and improve techniques in text classification, sentiment analysis, and feature extraction, significantly contributing to advancements in automated spam detection systems.

\subsection{The Neural Network Architectures}
\subsubsection{Feed Forward Neural Network}
The neural network architecture without dropout consists of a sequential arrangement of fully connected layers and nonlinear activation functions. Given an input vector \( \mathbf{x} \in \mathbb{R}^{2000} \), the network maps this input to an output vector \( \mathbf{y} \in \mathbb{R}^{20} \) representing the class scores. The mathematical formulation of the network operations is as follows:
\begin{align*}
\mathbf{h}_1 &= \text{ReLU}(\mathbf{W}_1 \mathbf{x} + \mathbf{b}_1), \\
\mathbf{h}_2 &= \text{ReLU}(\mathbf{W}_2 \mathbf{h}_1 + \mathbf{b}_2), \\
\mathbf{y} &= \mathbf{W}_3 \mathbf{h}_2 + \mathbf{b}_3,
\end{align*}
where \( \mathbf{W}_1 \in \mathbb{R}^{1024 \times 2000} \), \( \mathbf{W}_2 \in \mathbb{R}^{512 \times 1024} \), and \( \mathbf{W}_3 \in \mathbb{R}^{20 \times 512} \) are weight matrices, and \( \mathbf{b}_1 \in \mathbb{R}^{1024} \), \( \mathbf{b}_2 \in \mathbb{R}^{512} \), \( \mathbf{b}_3 \in \mathbb{R}^{20} \) are bias vectors.

\paragraph{Architecture with Dropout}
\vspace{0.3cm}
The architecture with dropout includes dropout layers after each ReLU activation in the hidden layers. This modification aims to randomly zero out a fraction of the outputs from the previous layer during training, which helps in preventing overfitting. The architecture is defined as follows:
\begin{align*}
\mathbf{h}_1 &= \text{ReLU}(\mathbf{W}_1 \mathbf{x} + \mathbf{b}_1), \\
\mathbf{h}_1' &= \text{Dropout}(\mathbf{h}_1, p=0.5), \\
\mathbf{h}_2 &= \text{ReLU}(\mathbf{W}_2 \mathbf{h}_1' + \mathbf{b}_2), \\
\mathbf{h}_2' &= \text{Dropout}(\mathbf{h}_2, p=0.5), \\
\mathbf{y} &= \mathbf{W}_3 \mathbf{h}_2' + \mathbf{b}_3,
\end{align*}
where \( p \) is the dropout probability, indicating the fraction of the input units to drop.

\subsubsection{Recurrent Neural Network}

The Recurrent Neural Network without dropout is streamlined, focusing on the essential properties of an RNN followed by a linear layer for classification:
\begin{align*}
\textbf{RNN Layer:} & \\
\mathbf{H}, \mathbf{h}_{\text{final}} &= \text{RNN}(\mathbf{X}, \mathbf{h}_0) \\
\textbf{Linear Output Layer:} & \\
\mathbf{y} &= \mathbf{W}_{\text{fc}} \mathbf{h}_{\text{final}} + \mathbf{b}_{\text{fc}}
\end{align*}
Where $\mathbf{X}$ is the input sequence, $\mathbf{H}$ denotes the hidden states, $\mathbf{h}_{\text{final}}$ is the final hidden state, and $\mathbf{W}_{\text{fc}}$, $\mathbf{b}_{\text{fc}}$ are the parameters of the fully connected output layer.

\paragraph{Architecture with Dropout}
\vspace{0.3cm} The architecture with dropout includes dropout layers to reduce overfitting by randomly dropping units during training:
\begin{align*}
\textbf{RNN Layer with Dropout:} & \\
\mathbf{H}, \mathbf{h}_{\text{final}} &= \text{RNN}(\mathbf{X}, \mathbf{h}_0; \text{dropout}=p) \\
\textbf{Dropout Before Linear Output Layer:} & \\
\mathbf{h}_{\text{final}'} &= \text{Dropout}(\mathbf{h}_{\text{final}}, p) \\
\textbf{Linear Output Layer:} & \\
\mathbf{y} &= \mathbf{W}_{\text{fc}} \mathbf{h}_{\text{final}'} + \mathbf{b}_{\text{fc}}
\end{align*}
Here, dropout is applied after the RNN layer and just before the linear output layer, introducing randomness that helps the model generalize better to unseen data.

\subsection{Results}

\begin{table}[h!]
\centering
\caption{Summary of Results for 20 News Group Dataset}
\begin{tabular}{|c|c|c|c|c|}
\hline
\textbf{Model Type} & \textbf{Configuration} & \textbf{Final Loss} & \textbf{Accuracy} & \textbf{Weighted F1-Score} \\ \hline
\multirow{2}{*}{Feed Forward Network} & Without Dropout & 0.0137 & 80.29\% & 0.8031 \\ \cline{2-5} 
                                      & With Dropout    & 0.0041 & 82.12\% & 0.8209 \\ \hline
\multirow{2}{*}{Recurrent Neural Network} & Without Dropout & 0.0286 & 80.13\% & 0.8115 \\ \cline{2-5} 
                                           & With Dropout    & 0.2821 & 81.41\% & 0.8200 \\ \hline
\end{tabular}
\end{table}

\begin{table}[h!]
\centering
\caption{Summary of Results for SMS Spam Dataset}
\begin{tabular}{|c|c|c|c|c|}
\hline
\textbf{Model Type} & \textbf{Configuration} & \textbf{Final Loss} & \textbf{Accuracy} & \textbf{Weighted F1-Score} \\ \hline
\multirow{2}{*}{Feed Forward Network} & Without Dropout & 0.0037 & 98.83\% & 0.9883 \\ \cline{2-5} 
                                      & With Dropout    & 0.0041 & 99.10\% & 0.9910 \\ \hline
\multirow{2}{*}{Recurrent Neural Network} & Without Dropout & 0.0059 & 98.92\% & 0.9883 \\ \cline{2-5} 
                                           & With Dropout    & 0.0059 & 99.19\% & 0.9918 \\ \hline
\end{tabular}
\end{table}

\begin{figure}[h!]
    \centering
    \includegraphics[width=0.8\textwidth]{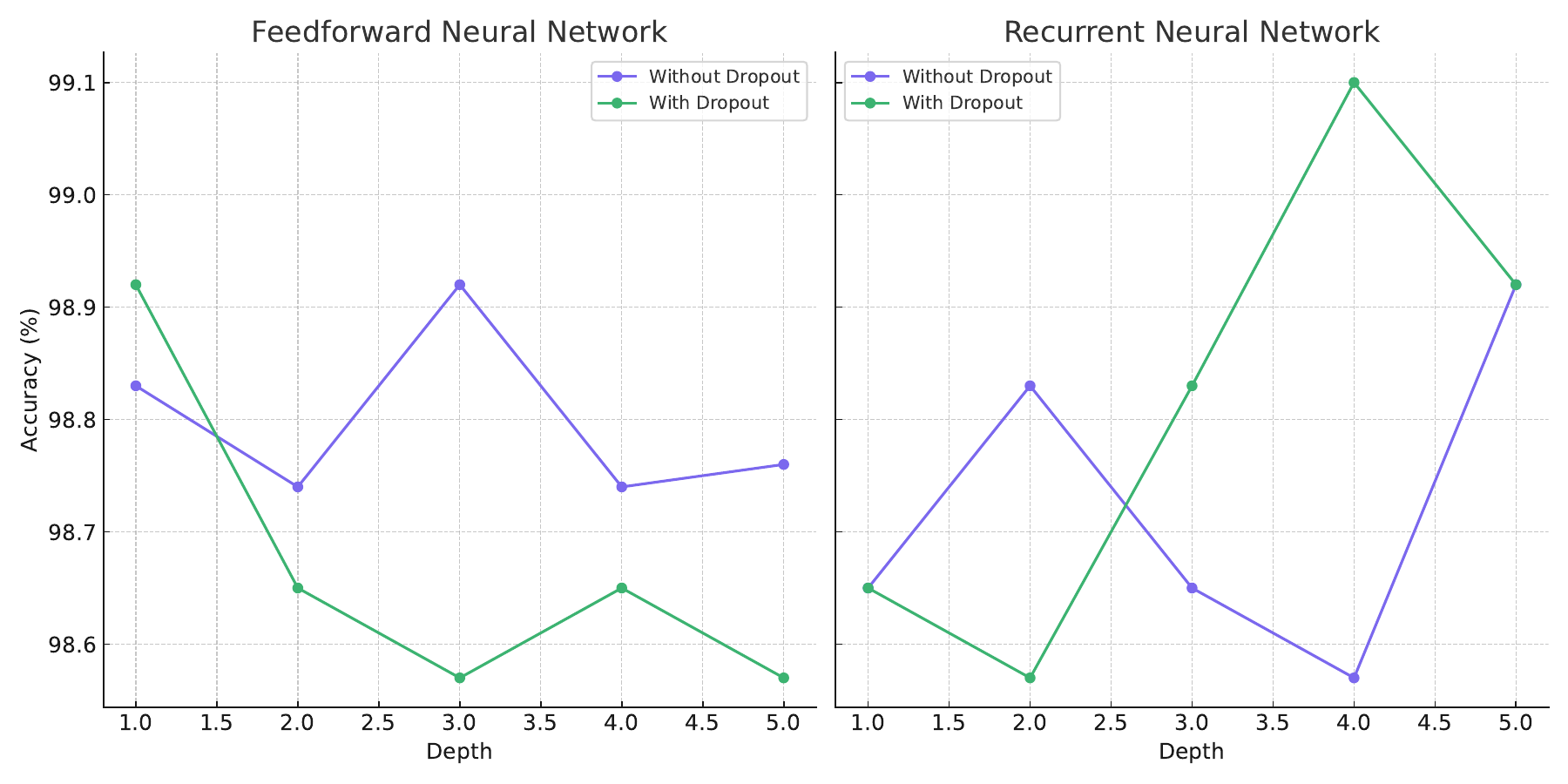}
    \caption{Accuracy comparison across depths for models with and without dropout on SMS Spam Dataset using TANNs.}
\end{figure}

\begin{figure}[h!]
    \centering
    \includegraphics[width=0.8\textwidth]{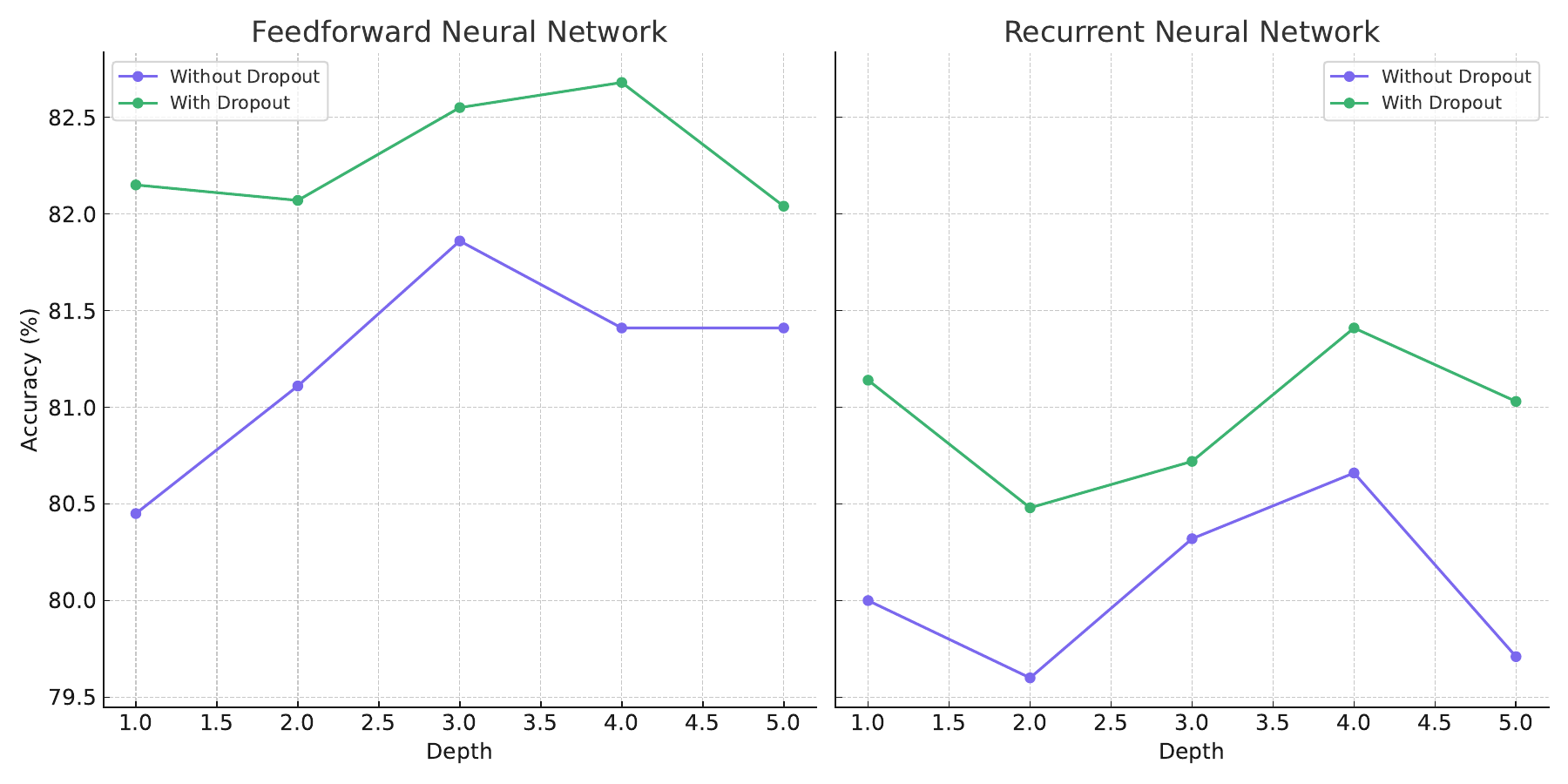}
    \caption{Accuracy comparison across depths for models with and without dropout on 20 NewsGroup Dataset using TANNs.}
\end{figure}

\begin{table}[h!]
\centering
\begin{tabular}{|>{\centering\arraybackslash}m{1.5cm}|>{\centering\arraybackslash}m{1.5cm}|>{\centering\arraybackslash}m{1.5cm}|>{\centering\arraybackslash}m{1.5cm}|>{\centering\arraybackslash}m{1.5cm}|>{\centering\arraybackslash}m{1.5cm}|}
\hline
\textbf{Depth} & \textbf{Metric} & \textbf{FNN without Dropout} & \textbf{FNN with Dropout} & \textbf{RNN without Dropout} & \textbf{RNN with Dropout} \\ \hline
\multirow{2}{*}{1} & Accuracy & 98.83\% & 98.92\% & 98.65\% & 98.65\% \\ \cline{2-6}
                   & F1-Score & 0.9562 & 0.9595 & 0.9498 & 0.9498 \\ \hline
\multirow{2}{*}{2} & Accuracy & 98.74\% & 98.65\% & 98.83\% & 98.57\% \\ \cline{2-6}
                   & F1-Score & 0.9533 & 0.9495 & 0.9559 & 0.9463 \\ \hline
\multirow{2}{*}{3} & Accuracy & 98.92\% & 98.57\% & 98.65\% & 98.83\% \\ \cline{2-6}
                   & F1-Score & 0.9595 & 0.9463 & 0.9498 & 0.9562 \\ \hline
\multirow{2}{*}{4} & Accuracy & 98.74\% & 98.65\% & 98.57\% & 99.10\% \\ \cline{2-6}
                   & F1-Score & 0.9530 & 0.9495 & 0.9467 & 0.9660 \\ \hline
\multirow{2}{*}{5} & Accuracy & 98.76\% & 98.57\% & 98.92\% & 98.92\% \\ \cline{2-6}
                   & F1-Score & 0.9534 & 0.9467 & 0.9595 & 0.9595 \\ \hline
\end{tabular}
\caption{Comparative performance of Feedforward and Recurrent Neural Networks by depth and dropout configuration on the SMS Spam Dataset}
\label{tab:network_performance_comparison}
\end{table}

\begin{table}[H]
\centering
\begin{tabular}{|>{\centering\arraybackslash}m{1.5cm}|>{\centering\arraybackslash}m{1.5cm}|>{\centering\arraybackslash}m{1.5cm}|>{\centering\arraybackslash}m{1.5cm}|>{\centering\arraybackslash}m{1.5cm}|>{\centering\arraybackslash}m{1.5cm}|}
\hline
\textbf{Depth} & \textbf{Metric} & \textbf{FNN without Dropout} & \textbf{FNN with Dropout} & \textbf{RNN without Dropout} & \textbf{RNN with Dropout} \\ \hline
\multirow{2}{*}{1} & Accuracy & 80.45\% & 82.15\% & 80.00\% & 81.14\% \\ \cline{2-6}
                   & F1-Score  & 0.80 & 0.82 & 0.80 & 0.81 \\ \hline
\multirow{2}{*}{2} & Accuracy & 81.11\% & 82.07\% & 79.60\% & 80.48\% \\ \cline{2-6}
                   & F1-Score  & 0.81 & 0.82 & 0.79 & 0.80 \\ \hline
\multirow{2}{*}{3} & Accuracy & 81.86\% & 82.55\% & 80.32\% & 80.72\% \\ \cline{2-6}
                   & F1-Score & 0.82 & 0.83 & 0.80 & 0.81 \\ \hline
\multirow{2}{*}{4} & Accuracy & 81.41\% & 82.68\% & 80.66\% & 81.41\% \\ \cline{2-6}
                   & F1-Score & 0.81 & 0.82 & 0.80 & 0.81 \\ \hline
\multirow{2}{*}{5} & Accuracy & 81.41\% & 82.04\% & 79.71\% & 81.03\% \\ \cline{2-6}
                   & F1-Score & 0.81 & 0.82 & 0.80 & 0.81 \\ \hline
\end{tabular}
\caption{Comparative performance of Feedforward and Recurrent Neural Networks by depth and dropout configuration on the 20 NewsGroup Dataset}
\label{tab:network_performance_comparison}
\end{table}

\subsection{Discussion}
Trie-Augmented Neural Networks (TANNs) introduce a pioneering approach by integrating trie structures with neural network elements, making significant strides in managing hierarchical and structured data. This unique combination allows for decomposing complex, large-scale neural networks into smaller, manageable units spread across the trie nodes, enhancing the efficiency of learning pertinent features by effectively segmenting the input space.

\subsubsection{Challenges in Training}
This design potentially complicates the training process, as it requires the model to not only learn the appropriate features from the data but also how to effectively segment the input space hierarchically. This segmentation might not align perfectly with the underlying patterns in the training data, potentially leading to inefficiencies in learning and higher initial loss values.

However, the successful deployment of TANNs hinges crucially on their training dynamics. Properly crafted segmentation strategies are imperative, as any shortcomings in how input spaces are divided could result in suboptimal performance. Additionally, the integration of complex neural networks within trie nodes could potentially escalate computational demands and resources, potentially making TANN models cumbersome to train without substantial computational support.

\subsubsection{Segmentation Strategies}
It is crucial to recognize that precise segmentation of the input space is essential in Trie-Augmented Neural Networks (TANNs). An ineffective segmentation strategy could inadvertently separate critical features of the neural network, potentially impairing the learning capabilities of TANNs. To mitigate this risk, several strategies can be explored, including feature sharing, where common features are accessible across multiple trie nodes; balanced segmentation, which ensures that data is evenly distributed across the trie to prevent overfitting or underfitting in any single branch; hierarchical feature fusion, which combines features from different levels of the trie to enhance the learning context; and regularization techniques, which help to avoid overfitting by penalizing overly complex models. Furthermore, integrating hybrid architectures that combine trie-based and traditional neural network elements could provide a balance between localized and global feature processing. Empirical evaluations are crucial to understand the impact of these strategies on feature dilution and to ensure that TANNs can fully realize their potential in boosting neural network performance.

\subsubsection{Feature Isolation and Data Sparsity}
A notable challenge in the implementation of TANNs is that, unlike traditional neural networks where features learned at initial layers are propagated and refined through subsequent layers, the trie structure can lead to the isolation of certain features within specific branches. This isolation might prevent these features from contributing effectively to the overall learning process, thereby limiting the network's ability to build on foundational knowledge.

Another concern arises from the potential sparsity of data within individual trie nodes. Especially in deeper or more branched trie structures, nodes may end up processing only a small subset of the overall data, leading to a sparse distribution of information. This sparsity can severely hamper the ability of the neural networks within each node to learn robust, generalizable features, particularly in the initial layers where data is less processed and more prone to exhibiting sparsity-related issues.

\subsubsection{Mathematical Perspective}
Incorporating a mathematical perspective can further elucidate the challenges faced by Trie-Augmented Neural Networks (TANNs) compared to traditional AlexNet. Each node in TANN's trie structure represents a mini neural network, which contributes to an overall model complexity \(O(N \cdot M)\), where \(N\) is the number of nodes in the trie and \(M\) represents the complexity of each mini neural network. This complexity not only increases the parameter space but also affects the gradient descent process during backpropagation.

The loss function gradient \(\nabla L\) for TANNs is computed not just over a single network but across multiple interconnected networks. This can be represented mathematically as:

\[
\nabla L = \sum_{i=1}^{N} \nabla L_{i} \cdot W_{i}
\]

Here, \(\nabla L_{i}\) is the gradient of the loss function for the \(i\)-th mini neural network, and \(W_{i}\) is the weight matrix associated with the \(i\)-th node. This sum indicates that the gradients from individual networks are influenced by their respective positions and connections within the trie, which can lead to complex gradient flow patterns. Moreover, if the weight updates \(\Delta W_{i}\) for each mini-network are not properly regulated, it could lead to issues such as gradient vanishing or exploding, especially given the compounded effect of multiple networks. This complexity underscores the need for precise control mechanisms, such as adaptive learning rates or advanced regularization techniques, to ensure stable training dynamics across the heterogeneous structure of TANNs.

\subsubsection{Performance in Specific Tasks}
TANNs excel in environments with clearly defined features, as evidenced by their performance on binary classification tasks such as XOR/AND, OR, and spam detection, where they achieve accuracy and F1-scores nearing perfection. Their moderate success in adapting to varied text classification tasks using Feed Forward Networks also showcases their potential versatility across different data types.

We conducted experiments with trie depths varying from 1 to 5. Our observations revealed that while increasing the trie depth initially improved performance, exceeding a certain threshold led to deterioration. This suggests that there is an optimal trie depth for the TANN structure under specific configurations. However, variations in hyperparameters, training strategies, and implementation details could potentially shift this optimal depth.

Moreover, the implementation of dropout in both Feed Forward and Recurrent Neural Networks using TANNs on the 20 News Group dataset shows that dropout effectively combats overfitting, a common issue in complex models such as TANNs. This adaptation is crucial in maintaining robustness, especially when facing diverse and less structured datasets like the 20 News Group, where accuracy lags behind that of more straightforward datasets like SMS Spam.

\subsubsection{Remarks}
Despite these challenges, TANNs demonstrate significant potential for enhancing performance in simpler models, particularly if the trie depth and hyperparameters are carefully optimized. However, their effectiveness in complex models remains constrained by the current early stages of their development. The architecture's novelty and the preliminary nature of its training and inference processes call for extensive further research and refinement. Continued experimentation and development are required to optimize TANNs' configuration and fully leverage their unique capabilities.

\section{Limitations of Trie-Augmented Neural Networks (TANNs)}

The computational cost of Trie-Augmented Neural Networks (TANNs) varies significantly based on multiple factors including the complexity of training logic, the architecture of the neural network within each node, and the depth of the trie data structure. Choosing between simpler networks with fewer layers in each node's network and a shallow trie depth can significantly reduce computational and memory demands. Conversely, employing a convolutional neural network (CNN) with multiple layers at each node and a deep trie structure increases both computational cost and memory usage. Balancing these factors is crucial for optimal performance and resource efficiency in TANNs.

\subsection{Depth of Trie}
The trie's depth is pivotal during initialization. A shallow trie may fail to capture the complexities of the data, leading to potential underperformance. On the other hand, an excessively deep trie can render the TANN inefficient and redundant due to increased computational and memory demands. Thus, it is essential to balance trie depth and network complexity within each node to optimize performance and resource use. Future developments could explore a self-adjusting trie that modifies its depth based on the training data, alongside node networks that adjust based on optimization variables.

\subsection{Computational Complexity}
The computational complexity of TANNs is influenced by two primary factors: the structure of the trie and the complexity of the neural networks embedded at each node. For a balanced trie, the complexity is represented as \(O(h)\), where \(h\) is the trie's height. This complexity arises from navigating from the root to a leaf node during each operation. In contrast, an unbalanced trie presents a less predictable complexity, with the worst-case scenario expressed as \(O(\text{max}(h))\), indicating the maximum height within the trie.

This complexity is further compounded by the computational demands of the neural networks at each node, which can be quantified by a multiplier, \(t\), representing the time spent during each pass at each node. Factors influencing \(t\) include the neural network architecture, activation functions, batch size, optimization algorithm, and input data dimensionality at each node. An approximation for \(t\) could be represented as \(t \approx N \times L \times C\), where \(N\) is the number of neurons, \(L\) the number of layers, and \(C\) the average computational cost per neuron per layer.

Consequently, the overall complexity for a balanced Trie can be modeled as \(O(h \cdot t)\), and for an unbalanced Trie as \(O(\text{max}(h) \cdot t)\). These considerations underline the importance of carefully selecting neural network architectures and optimization strategies at each node to manage the \(t\) value effectively, thus impacting the overall computational complexity and efficiency of TANNs.

\section{Further Research}

This architecture holds potential by augmenting neural networks' ability to segment the input space by breaking it into smaller segments, with each node specializing in a specific aspect of the process. This holds potential in hierarchical decision-making, recommender systems, text classification, and medical diagnosis, in which interpretability is essential. As each node within the trie focuses on a particular subset of the data, the network can provide more targeted and understandable outputs. This is crucial in applications where decision paths must be transparent and justifiable. The paper only introduced the architecture skeleton; with limited testing on domain specific tasks in text classification, further experimentation and refining would be crucial to solving more extensive scale and nuanced problems not tackled in this paper. Future research should explore optimal node architectures and trie structure configurations to enhance performance and efficiency. Additionally, real-world applications and datasets can provide practical insights and challenges that would help fine-tune the model parameters and improve the generalization capabilities of Trie Augmented Neural Networks. This next development phase is essential to moving from a theoretical framework to a robust, deployable model that can meet the complex demands of modern machine learning tasks.

\section{Conclusion}

This paper introduces a novel neural network architecture that effectively segments the input space across nodes. This architecture has demonstrated promising results in benchmarks such as the XOR and AND/OR logic gate problems, as well as in text classification tasks using the 20 NewsGroup and SMS Spam Collection datasets. The paper describes the foundational architecture and identifies several critical areas requiring further refinement to unlock the full potential of this approach. These areas include optimizing the trie depth, reducing computational complexity, and enhancing training processes. Moreover, it discusses strategies to mitigate overfitting and enhance the generalizability of the model. The primary goal of this work is to present this innovative architecture and stimulate further research in this area.

\section*{References}

\begin{itemize}[left=0pt, label={}]
    \item Arjovsky, M., Chintala, S., Bottou, L. Wasserstein GAN. arXiv preprint arXiv:1701.07875 (2017).
    \item Arora, S., Ge, R., Liang, Y., Ma, T., Zhang, Y. Generalization and equilibrium in generative adversarial nets (GANs). arXiv preprint arXiv:1703.00573 (2017).
    \item Beaulieu-Jones, B.K., Wu, Z.S., Williams, C., Greene, C.S. Privacy-preserving generative deep neural networks support clinical data sharing. bioRxiv (2017), 159756.
    \item Bengio, Y., Thibodeau-Laufer, E., Alain, G., Yosinski, J. Deep generative stochastic networks trainable by backprop. In ICML’2014 (2014).
    \item Brundage, M., Avin, S., Clark, J., Toner, H., Eckersley, P., Garfinkel, B., Dafoe, A., Scharre, P., Zeitzoff, T., Filar, B., Anderson, H., Roff, H., Allen, G.C., Steinhardt, J., Flynn, C., hÉigeartaigh, S.Ó., Beard, S., Belfield, H., Farquhar, S., Lyle, C., Crootof, R., Evans, O., Page, M., Bryson, J., Yampolskiy, R., Amodei, D. The Malicious Use of Artificial Intelligence: Forecasting, Prevention, and Mitigation. ArXiv e-prints (Feb. 2018).
    \item Danihelka, I., Lakshminarayanan, B., Uria, B., Wierstra, D., Dayan, P. Comparison of maximum likelihood and GAN-based training of real nvps. arXiv preprint arXiv:1705.05263 (2017).
    \item Goodfellow, I., Pouget-Abadie, J., Mirza, M., Xu, B., Warde-Farley, D., Ozair, S., Courville, A., Bengio, Y. Generative adversarial nets. In Advances in Neural Information Processing Systems (2014).
    \item He, K., Zhang, X., Ren, S., Sun, J. Deep residual learning for image recognition. In Proceedings of the IEEE conference on computer vision and pattern recognition (2016).
    \item Hochreiter, S., Schmidhuber, J. Long short-term memory. Neural Computation, 9(8), 1735-1780 (1997).
    \item Kingma, D.P., Welling, M. Auto-Encoding Variational Bayes. arXiv preprint arXiv:1312.6114 (2013).
    \item Krizhevsky, A., Sutskever, I., Hinton, G.E. ImageNet classification with deep convolutional neural networks. In Advances in Neural Information Processing Systems (2012).
    \item Szegedy, C., Liu, W., Jia, Y., Sermanet, P., Reed, S., Anguelov, D., Erhan, D., Vanhoucke, V., Rabinovich, A. Going deeper with convolutions. In Proceedings of the IEEE conference on computer vision and pattern recognition (2015).
    \item Vaswani, A., Shazeer, N., Parmar, N., Uszkoreit, J., Jones, L., Gomez, A.N., Kaiser, L., Polosukhin, I. Attention is all you need in Advances in neural information processing systems (2017).
    \item Radford, A., Narasimhan, K., Salimans, T., Sutskever, I. Improving language understanding by generative pre-training. OpenAI preprint (2018).
    \item Devlin, J., Chang, M.W., Lee, K., Toutanova, K. BERT: Pre-training of deep bidirectional transformers for language understanding. arXiv preprint arXiv:1810.04805 (2018).
    \item Silver, D., Huang, A., Maddison, C.J., Guez, A., Sifre, L., Van Den Driessche, G., Schrittwieser, J., Antonoglou, I., Panneershelvam, V., Lanctot, M., Dieleman, S. Mastering the game of Go with deep neural networks and tree search: nature, 529(7587), 484–489 (2016).
    \item Williams, R.J., Zipser, D. A learning algorithm for continually running fully recurrent neural networks. Neural Computation, 1(2), 270–280 (1989).
    \item Bengio, Y., Courville, A., Vincent, P. Representation learning: A review and new perspectives. IEEE transactions on pattern analysis and machine intelligence, 35(8), 1798-1828 (2013).
    \item Ronneberger, O., Fischer, P., Brox, T. U-Net: Convolutional networks for biomedical image segmentation. In International Conference on Medical image computing and computer-assisted intervention (2015).
    \item Dosovitskiy, A., Beyer, L., Kolesnikov, A., Weissenborn, D., Zhai, X., Unterthiner, T., Dehghani, M., Minderer, M., Heigold, G., Gelly, S., Uszkoreit, J., Houlsby, N. An image is worth 16x16 words: Transformers for image recognition at scale—arXiv preprint arXiv:2010.11929 (2020).
    \item Li, Y., Wang, N., Shi, J., Hou, X., Liu, J. Adaptive batch normalization for practical domain adaptation. Pattern Recognition, 80, 109–117 (2018).
    \item Bahdanau, D., Cho, K., Bengio, Y. Neural machine translation by jointly learning to align and translate. arXiv preprint arXiv:1409.0473 (2014).
    \item Mnih, V., Kavukcuoglu, K., Silver, D., Graves, A., Antonoglou, I., Wierstra, D., Riedmiller, M. Human-level control through deep reinforcement learning: nature, 518(7540), 529–533 (2015).
    \item Hinton, G.E., Srivastava, N., Krizhevsky, A., Sutskever, I., Salakhutdinov, R. Improving neural networks by preventing co-adaptation of feature detectors. arXiv preprint arXiv:1207.0580 (2012).
    \item Sutskever, I., Vinyals, O., Le, Q.V. Sequence to sequence learning with neural networks. In Advances in Neural Information Processing Systems (2014).
    \item Goodfellow, I.J., Warde-Farley, D., Mirza, M., Courville, A., Bengio, Y. Maxout networks. In ICML (2013).
    \item Hochreiter, S., Bengio, Y., Frasconi, P., Schmidhuber, J. Gradient flow in recurrent nets: the difficulty of learning long-term dependencies. In A Field Guide to Dynamical Recurrent Neural Networks (2001).
    \item Krizhevsky, A., Hinton, G. Learning multiple features layers from tiny images: Technical report, University of Toronto (2009).
    \item Srivastava, N., Hinton, G., Krizhevsky, A., Sutskever, I., Salakhutdinov, R. Dropout: A simple way to prevent overfit neural networks. Journal of Machine Learning Research, 15(1), 1929-1958 (2014).
    \item McClelland, J.L., Rumelhart, D.E. Parallel distributed processing: Explorations in the microstructure of cognition. Volume 2: Psychological and biological models. MIT Press (1986).
\end{itemize}
\end{document}